\documentclass[conference]{IEEEtran}
\IEEEoverridecommandlockouts
\usepackage{cite} \usepackage{amsmath,amssymb,amsfonts}
\usepackage{algorithmic} \usepackage{graphicx} \usepackage{textcomp}
\usepackage{xcolor} \usepackage{hyperref} \usepackage[normalem]{ulem}

\def\BibTeX{{\rm B\kern-.05em{\sc i\kern-.025em b}\kern-.08em
    T\kern-.1667em\lower.7ex\hbox{E}\kern-.125emX}}

\usepackage{fancyhdr}
\begin{document}

\title{Emotion Twenty Questions Dialog System for Lexical Emotional
  Intelligence \thanks{ We thank the following for support and
    funding: University of St. Thomas Graduate Programs in Software's
    Center for Applied Artificial Intelligence and University of
    St. Thomas Center for Faculty Development Graduate Research Team
    Grant.  }  }

\author{
\IEEEauthorblockN{Abe Kazemzadeh}
\IEEEauthorblockA{\textit{Graduate Programs in Software} \\
\textit{University of St. Thomas}\\
St. Paul, MN, U.S.A. \\
orcid:0000-0002-1851-294X}
\and
\IEEEauthorblockN{ Adedamola Sanusi}
\IEEEauthorblockA{\textit{Graduate Programs in Software} \\
\textit{University of St. Thomas}\\
St. Paul, MN, U.S.A. \\
ola.sanusi@stthomas.edu}
\and
\IEEEauthorblockN{Huihui (Summer) Nie}
\IEEEauthorblockA{\textit{Graduate Programs in Software} \\
\textit{University of St. Thomas}\\
St. Paul, MN, U.S.A. \\
summer.nie@stthomas.edu}

}

\maketitle
\thispagestyle{fancy}

\begin{abstract}
  
  This paper presents a web-based demonstration of Emotion Twenty
  Questions (EMO20Q), a dialog game whose purpose is to study how
  people describe emotions.  EMO20Q can also be used to develop
  artificially intelligent
  dialog agents that can play the game.  In
  previous work, an EMO20Q agent used a sequential Bayesian machine
  learning model and could play the question-asking role.  Newer
  transformer-based neural machine learning models have made it
  possible to develop an agent for the question-answering role.


  This demo paper describes the recent developments in the
  question-answering role of the EMO20Q game, which requires the agent
  to respond to more open-ended inputs.  Furthermore, we also describe
  the design of the system, including the web-based front-end, agent
  architecture and programming, and updates to earlier software used.


  The demo system will be available to collect pilot data during
  the ACII conference and this data will be used to inform future
  experiments and system design.



  An example of the demo system can be seen in
  Fig.~\ref{fig:emo20q-flask-websocket} and the live system can be
  accessed at \href{https://emo20q.org}{emo20q.org}.
\end{abstract}

\begin{IEEEkeywords}
emotions, natural language processing, dialog systems,
question-answering, EMO20Q, lexicon
\end{IEEEkeywords}

\begin{figure}[h]
\centering
\includegraphics[width=0.5\textwidth]{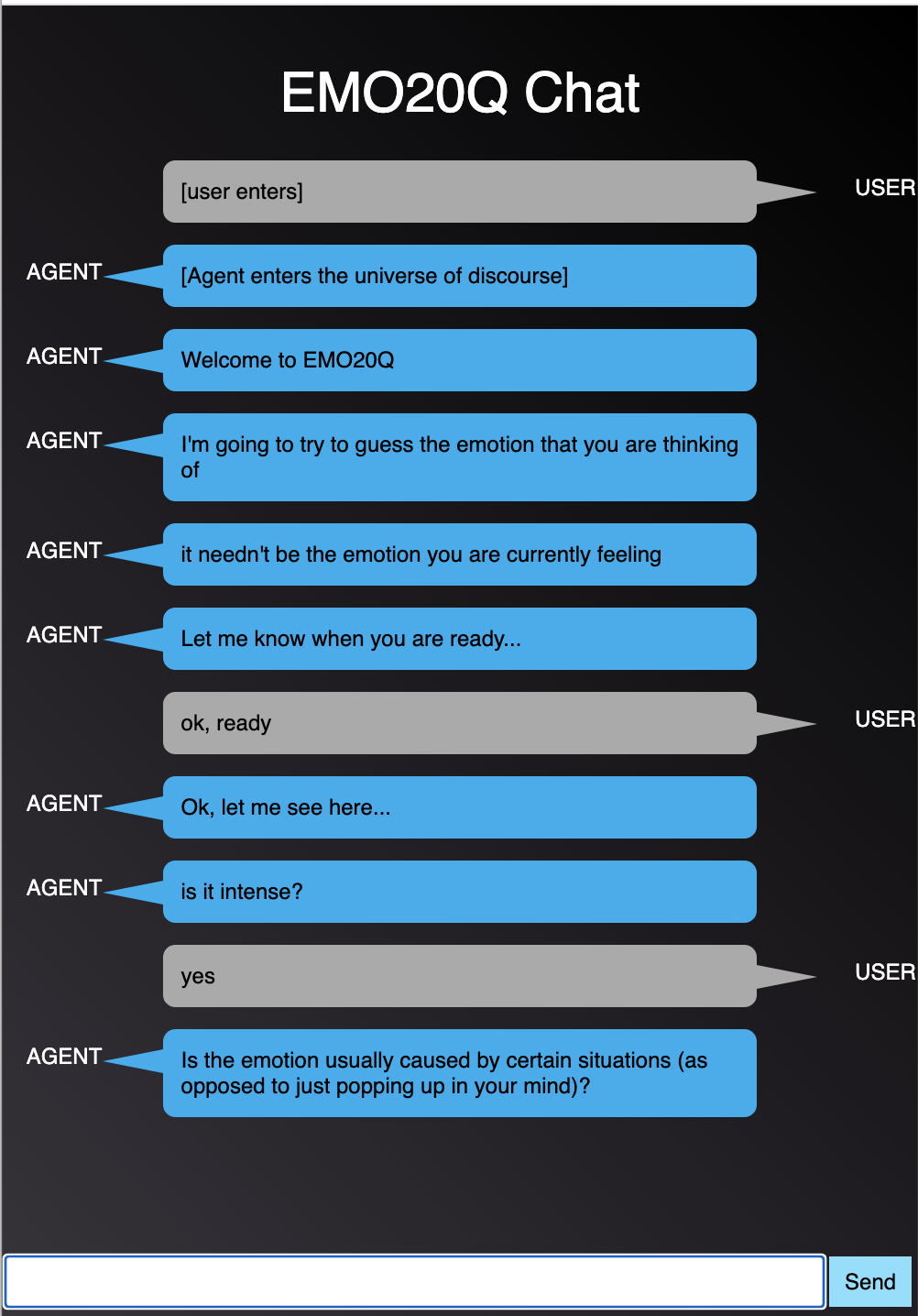}
\caption{Example of the web-based front end to the EMO20Q dialog
  system.}
\label{fig:emo20q-flask-websocket}
\end{figure}

\section{Introduction}

This demo system aims to use a dialog agent to collect question-answer
data about human emotions.  We use the emotion twenty questions game
(EMO20Q) as the experimental setting.  In EMO20Q, one player picks an
emotion word and the other player tries to guess it in twenty or fewer
turns. The player that picks the emotion word plays the {\it
  question-answering} role, because they answer questions about the
emotion word that they picked.  The player who tries to guess the
emotion word plays the {\it question-asking} role, because they ask
questions in order to identify the emotion word.

Both player roles for EMO20Q could be played by humans or computers.
Past research collected data of humans playing both roles and then
used this data to train dialog agents to play the question-asking role
\cite{Kazemzadeh2012}.  This past work used a sequential Bayesian
probabilistic model for the question-asking role in the game
\cite{Kazemzadeh2012}: the input was a sequence of question-answer
pairs. The question set comes from human-human dialogs and the
particular question that the agent asks is chosen based on the dialog
history. The answers come from the user's corresponding responses.
Posterior probabilities of the emotion words are updated based on
these sequential question-answer inputs.

The current work aims to extend the approach to a dialog agent that
can play the question-answering role. In this case, the agent picks
the emotion word from words seen in previously collected data (167
words) and answers questions posed by the user. In this case, the
agent's inputs are very open-ended: any possible yes/no questions
about emotions.  Many questions, especially earlier in the game, have
been seen before (e.g., ``is it a positive emotion?'', ``is it a
negative emotion?'', ``is it an emotion that is directed at another
person?'', ``is it an emotion that lasts a long time?''), but in
principle, any questions could be asked.  In comparison, the inputs to
the question-asking role are more limited.  The question-asking agent
asks questions from the set of questions seen in human-human data and
its inputs are answers to these yes-no questions. The answers are not
restricted to yes or no, but it is fairly trivial to bucket these
responses into yes/no/other categories.

To deal with the open-ended question-answering task, we aim to
leverage neural large language models (LLM), which encode linguistic
knowledge from large pre-training datasets into neural networks,
which can then be fine-tuned into task-specific models using relatively
smaller datasets.  

The goal for EMO20Q dialog system is to use dialog agents to test
hypotheses about how well automated systems can understand language
about emotions.



\section{Web Front End}

To simply display the front end of the demo system, we used Flask, a
lightweight Python web framework.  Using the simple request-response
interaction of a basic web server would have been an option, but it
would require a page reload for every dialog turn and could result in
slow response times, especially in cases where prediction requires
beam search (the question-answering role).  To prevent page reloads
and response delays, we used WebSockets via the Socket.io library,
which enables bidirectional communication between the browser and
web server.  Dialog events, i.e. turns from the user (via the browser)
or agent (via the server), trigger updates to a typical speech bubble
dialog display, as seen in Fig.~\ref{fig:emo20q-flask-websocket}.

\section{Dialog Agent}

The dialog agent was trained using a method inspired by wizard of Oz
approaches \cite{Fraser1991} and games with a purpose
\cite{Ahn2004}. First, non-expert human players played both roles of
EMO20Q, which provided an initial set of training data. Then, we
created automated agent for the question-asking role by finding the
conditional probabilities of (question, answer) pairs given emotion
words and using these to create a sequential Bayesian probabilistic
model \cite{Kazemzadeh2012}, which provided a further source of
training data.  The current work has focused on the question-answering
role.  This current work has benefited from recent trends in
transfer-learned deep neural language models.  In particular
this demo uses BERT \cite{Devlin2019} to classify (emotion, question)
pairs into ``yes'', ``no'', and ``maybe'' categories. Although our
system achieves reasonable performance of about 72\% accuracy on
question-answering, we plan that the demo system will be an
experimental tool to more fully evaluate this and other models.

The programming of the agent is based on a generalized pushdown
automaton (GPDA) \cite{Allauzen2012}.  The abilities of a general
finite state automata are sufficient for most of the dialog behavior,
but the pushdown stack is used to maintain contextual information that
represents short-term/working memory in the question-asking role. The
transition graph of the automaton is implemented with the NetworkX Python
library.

\section{Discussion and Future Work}

This paper presented an overview of a tool that we will use to test
the abilities of automated systems to talk about emotions with human
users.  In this demo, we use BERT, an encoder-only model set up as a
classification model to provide yes/no/maybe answers.  This
classification approach fits with the existing approaches used by the
question-asking agent. Other transformer models that include decoders
may allow for more fluent output and may be able to encompass the
functionality currently covered by the dialog graph,  so we anticipate
that this will be an area of future testing.

Although the agent in our demo appears to the user as a single agent
playing both roles, the design of the functionality of this single
agent is currently implemented as two separate agents with different
machine learning models built on the same data. One area of future
research is to design a more unified approach where machine learning
models are shared between the two EMO20Q roles.  This approach could
enable adversarial training scenarios where two automated agents play
each other.



Another issue is the ordering of EMO20Q roles.  Currently, the demo
system makes the user play the question-answering role first, then the
question-asking role. This order may induce experimental effects.  For
example, playing as question-answerer first may prime the human user
to reuse the agent's questions.


In conclusion, we demonstrate an automated system that plays EMO20Q
using a web-based chat interface and a mix of older probabilistic
models for the question-asking role and a fine-tuned BERT model for
the question-answering role.


\section{Ethics Statement}

This research has been approved by the
University of St. Thomas
institutional review board (IRB).  Data is not collected with
personally identifiable information (PII) and the collected data will be
examined again for PII before public dissemination.

\bibliographystyle{ieeetr}

\end{document}